\documentclass[letterpaper]{article} 
\usepackage{aaai25}  
\usepackage{times}  
\usepackage{helvet}  
\usepackage{courier}  
\usepackage[hyphens]{url}  
\usepackage{graphicx} 
\usepackage[most]{tcolorbox}
\usepackage{natbib}  
\usepackage{caption} 
\usepackage{times}  
\usepackage{courier}  
\usepackage[hyphens]{url}  
\usepackage{graphicx} 
\usepackage{algorithm2e}
\usepackage{booktabs}
\usepackage{multirow}
\usepackage{slashbox}
\usepackage{amssymb} 
\usepackage{listings}
\usepackage{caption}
\usepackage{subcaption}
\usepackage{tikz}
\usepackage{comment}
\usepackage{tcolorbox} 

\usepackage{xcolor}

\usepackage{amsmath}   
\usepackage{amssymb}   
\usepackage{booktabs}  
\usepackage{graphicx}  
\usepackage{rotating}  
\newcommand{\rot}[1]{\rotatebox{0}{#1}} 

\definecolor{keywordcolor}{RGB}{0,0,139}
\definecolor{variablecolor}{RGB}{0,100,0}
\definecolor{green}{RGB}{0,255,0}
\definecolor{blue}{RGB}{0,0,255}
\definecolor{orange}{RGB}{255,165,0}
\definecolor{red}{RGB}{255,0,0}
\definecolor{purple}{RGB}{128,0,128}
\definecolor{cyan}{RGB}{0,255,255}
\definecolor{magenta}{RGB}{255,0,255}
\definecolor{yellow}{RGB}{255,255,0}
\definecolor{brown}{RGB}{139,69,19}
\definecolor{gray}{RGB}{128,128,128}
\definecolor{pink}{RGB}{255,182,193}
\definecolor{teal}{RGB}{0,128,128}
\definecolor{olive}{RGB}{128,128,0}
\definecolor{lightblue}{RGB}{173,216,230}
\definecolor{darkblue}{RGB}{0,0,139}

\lstdefinelanguage{mygrammar}{
  keywords={Asset, Class, Component, Name, Problem, Type, Data, Source},
  keywordstyle=\color{keywordcolor}\bfseries,
  moredelim=[s][\color{variablecolor}]{<}{>},
  moredelim=*[s][\color{black}]{::=},
}

\lstnewenvironment{mygrammar}{\lstset{language=mygrammar, basicstyle=\ttfamily}}{}

\newtcolorbox{mybox}[1]{
  enhanced,
  colback=lightblue!30,
  colframe=black,
  arc=3mm,
  fonttitle=\color{white}\bfseries,
  title=#1,
  coltitle=white,
  boxrule=0.5mm
}

\begin{document}

\title{LLM Assisted Anomaly Detection Service for Site Reliability Engineers: Enhancing Cloud Infrastructure Resilience}

\author{
    Nimesh Jha\textsuperscript{\rm 1},
    Shuxin Lin\textsuperscript{\rm 2},
    Srideepika Jayaraman\textsuperscript{\rm 2},
    Kyle Frohling\textsuperscript{\rm 3},
    Christodoulos Constantinides\textsuperscript{\rm 4},
    Dhaval Patel\textsuperscript{\rm 2}
    }
\affiliations{
    \textsuperscript{\rm 1}IBM Infrastructure,
    \textsuperscript{\rm 2}IBM Research, 
    \textsuperscript{\rm 3}IBM API Hub,
    \textsuperscript{\rm 4}IBM\\ 
    TJ Watson Research Center\\
    New York, NY 20004 USA\\
    {\{nimeshjha@in.,shuxin.lin@,j.srideepika@,frohling@us, christodoulos.constantinides@,pateldha@us.\}}ibm.com
}

\maketitle

\begin{abstract}
This paper introduces a scalable Anomaly Detection Service with a generalizable API tailored for industrial time-series data, designed to assist Site Reliability Engineers (SREs) in managing cloud infrastructure. The service enables efficient anomaly detection in complex data streams, supporting proactive identification and resolution of issues. Furthermore, it presents an innovative approach to anomaly modeling in cloud infrastructure by utilizing Large Language Models (LLMs) to understand key components, their failure modes, and behaviors. 

A suite of algorithms for detecting anomalies is offered in univariate and multivariate time series data, including regression-based, mixture-model-based, and semi-supervised approaches. We provide insights into the usage patterns of the service, with over 500 users and 200,000 API calls in a year. The service has been successfully applied in various industrial settings, including IoT-based AI applications. We have also evaluated our system on public anomaly benchmarks to show its effectiveness.

By leveraging it, SREs can proactively identify potential issues before they escalate, reducing downtime and improving response times to incidents, ultimately enhancing the overall customer experience.

We plan to extend the system to include time series foundation models, enabling zero-shot anomaly detection capabilities.
\end{abstract}
\maketitle

\section{Introduction}
The increasing adoption of cloud computing has led to a surge in demand for stability and reliability, making the role of Site Reliability Engineers (SREs) more critical than ever. To ensure seamless operations, SREs \cite{beyer2016sre} require automation systems that can streamline their daily activities and provide real-time visibility into the performance and health of cloud infrastructure. This enables them to identify potential issues before they escalate, reducing downtime and improving response times to incidents.

Currently, SREs utilize advanced monitoring dashboards to proactively monitor cloud infrastructure, but still face challenges in detecting and preventing incidents. When a service or application experiences an outage, SREs are forced to raise incidents, which can lead to a negative perception of the service provider's ability to detect and prevent incidents. To address this, an automated issue detection service or dashboard is essential, providing SREs with real-time insights into cloud infrastructure performance and health.

Our approach utilizes a Deep Learning-based Anomaly Detection approach to implement an example workflow, built on our publicly deployed Anomaly Detection Service API. This has significant advantages for cloud monitoring, enabling real-time identification of anomalies, thus improving reliability, optimizing performance, and ensuring compliance with service level agreements. This area has garnered significant attention from both professionals and researchers \cite{6753822, 9284273, 9402147, article}.

\begin{figure*}[h!]
    \centering
    \includegraphics[width=0.85\linewidth]{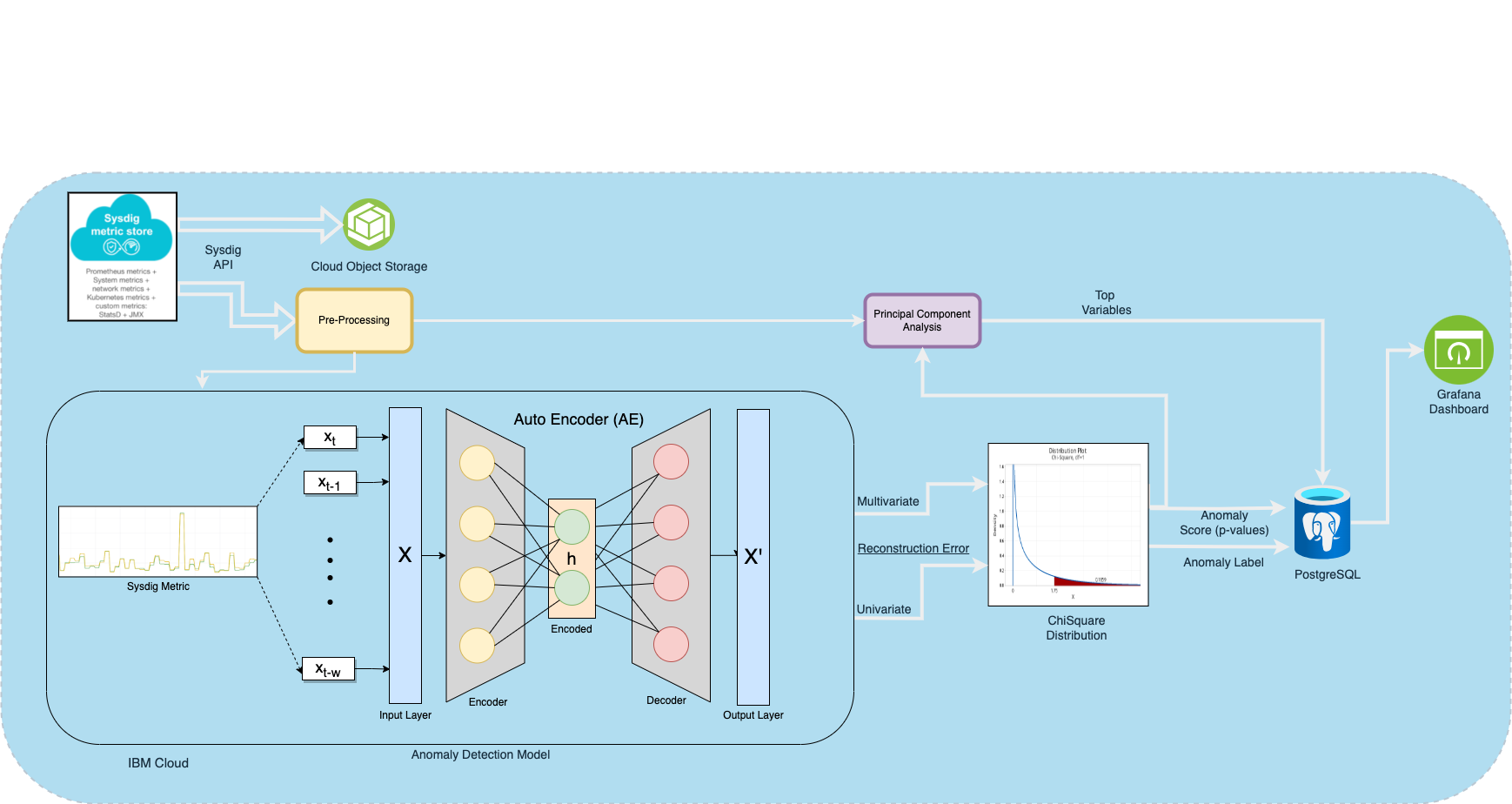}
    \caption{Overview of the end-to-end Cloud Metric Monitoring System using Anomaly Detection}
    \label{TA}
\end{figure*}

Our system is designed to provide a robust and scalable anomaly detection framework for cloud-based services, enabling the efficient identification of anomalies and ensuring the reliability of cloud infrastructure. We have developed a modular architecture that allows us to process large volumes of data efficiently and provide real-time insights into the health of cloud infrastructure, thereby facilitating prompt issue detection and resolution. We also explore the various personas who utilize our deployed system, including Site Reliability Engineers (SREs) and Cloud Services Developers. We highlight the features and functionalities of our system that cater to the specific needs of each persona, thus enhancing their productivity, efficiency, and overall user experience. Furthermore, we delve into the technical details of our underlying API, which powers our deployed system. This includes an overview of the API's design principles, implementation, and key features that enable real-time anomaly detection, predictive analytics, and data-driven decision-making. The API is deployed on the IBM Cloud infrastructure, which utilizes auto-scaling and load-balancing features to ensure efficient handling of dynamic workloads. Furthermore, the API's underlying architecture is designed to be data-agnostic, supporting a range of data types (i.e., univariate and multivariate time series, as well as tabular data) and enabling adaptability to various use cases. 
By providing a comprehensive understanding of our system's architecture, functionality, and technical details, we aim to demonstrate the effectiveness, reliability, and scalability of our anomaly detection framework, and its potential to transform cloud monitoring and management.
In the next section, we discuss how Large Language Models (LLMs) can assist in the selection of desired metrics for cloud infrastructure monitoring and in identifying the most relevant behavior of each metric to monitor.

\subsection{LLM-Assisted Anomaly Modelling}
\begin{figure}[!h]
\centering
\includegraphics[scale=0.5]{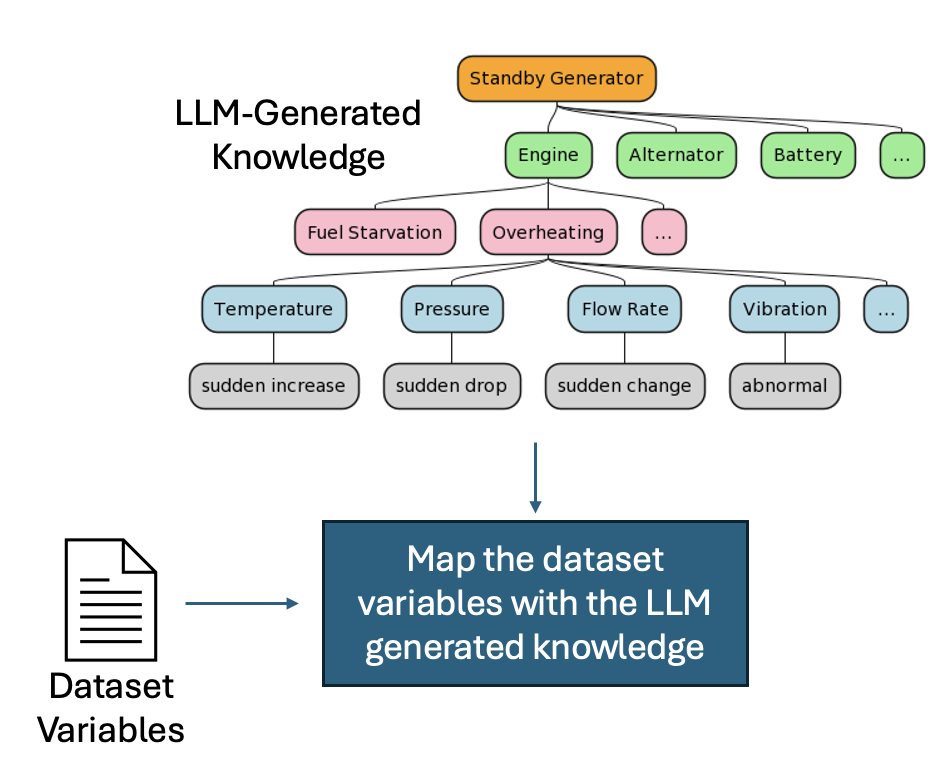} 
\caption{Mapping LLM-Generated Knowledge with the right dataset variables for building a more informed anomaly model for any industrial assets}
\label{casy}
\end{figure}

Cloud computer scientists aim to develop anomaly models that capture anomalous behaviors of various components within the cloud infrastructure. To achieve this goal, we systematically exploited pre-trained Large Language Models (LLMs) to generate diverse failure modes, their corresponding metrics for monitoring, and associated behavioral patterns. We then mapped the derived knowledge to relevant variables from the dataset. To illustrate this mapping concept, we selected the standby generator, a common industrial asset utilized in hospitals and data centers, as a specific use case (Figure \ref{casy}).

The workflow goes as follows:

\begin{enumerate}
  \item A pre-trained LLM like Llama \cite{touvron2023llama} or Mistral \cite{jiang2024mixtral} is prompted to identify the different components that exist in a cloud infrastructure
  \item For each component generated, it generates the different failure modes
  \item For each generated failure mode, it generates the metrics to monitor and what would constitute an anomalous behavior
  \item The LLM maps the generated knowledge of the failure modes, the relevant sensors and their behavior to the right cloud infrastructure metrics from our dataset 
\end{enumerate}

A cloud infrastructure consists of multiple components and subcomponents that each of them requires monitoring. 
A nonexhaustive list of components is as follows: 
\begin{itemize}
    \item Servers that consist of CPUs, Memory and Storage
    \item Network components which consist of Routers, Switches and Load Balancers
    \item Security Components which consist of Firewalls and 
    Intrusion Detection Systems 
    \item Power Systems that consist of Uninterruptible Power Supplies (UPS), Power Distribution Components like Switchgear and Transformers
\end{itemize}

Each component in the cloud infrastructure can fail for different reasons and has different metrics to monitor. For a detailed example, see Appendix~\ref{llm_examp}.

\section{Anomaly Detection System Workflow}
In this section, we first present a deep learning-based anomaly detection system for cloud monitoring, as illustrated in Figure \ref{TA}. The technical architecture outlines the flow of using Sysdig Metrics \cite{sysdig_docs} as input data, which is then processed by the Anomaly Detection API employing a \texttt{DNN\_AutoEncoder} based deep learning model, known as ReconstructAD. Additionally, the Chi-Square Distribution is utilized as a statistical method to extract p-values as anomaly scores and threshold values, enabling the determination of anomaly labels.
\begin{enumerate}
    \item \textbf{Data Capture}
The initial step in our system involves capturing data from IaaS (Infrastructure as a Service) multizone regions (MZR), which represent the cloud infrastructure resources, including virtual machines, network interfaces, and storage volumes. This data encompasses various metrics, such as:

\begin{enumerate}
\item Network traffic
\item Memory usage
\item Disk I/O
\item CPU utilization
\end{enumerate}

\item \textbf{Data Store}
The collected data is then stored in Cloud Object Storage for future analysis through a dedicated pipeline. Another pipeline is utilized to feed the data into a pre-processing stage, where the data is processed to render it consumable by the Anomaly Detection API.

\item \textbf{Anomaly Detection}
Autoencoder-based time series representation is an interesting research problem \cite{sym12081251,10.1609/aaai.v33i01.33011409,zong2018deep}.Our approach, the ReconstructAD algorithm
using the Anomaly Detection Service API\footnote{\url{https://developer.ibm.com/apis/catalog/ai4industry--anomaly-detection-product/}}, leverages a deep neural network \texttt{DNN\_AutoEncoder} \cite{aaai2022} based detection method to extract anomalies from the data. This approach identifies unusual patterns or behavior that may indicate a problem. Notably, a comparative analysis of different anomaly detection categories, including PredAD (an algorithm with a lookback window) and RelationshipAD (an algorithm that considers co-variance between variables), revealed that ReconstructAD outperformed the others in terms of accuracy for the IaaS dataset. The algorithms PredAD, ReconstructAD, and RelationshipAD are accessible through the Anomaly Detection Service API\footnote{\url{https://developer.ibm.com/apis/catalog/ai4industry--anomaly-detection-product/api/API--ai4industry--anomaly-detection-api#batch_uni}}.

\begin{figure}[h]
\centering
\includegraphics[width=1\linewidth]{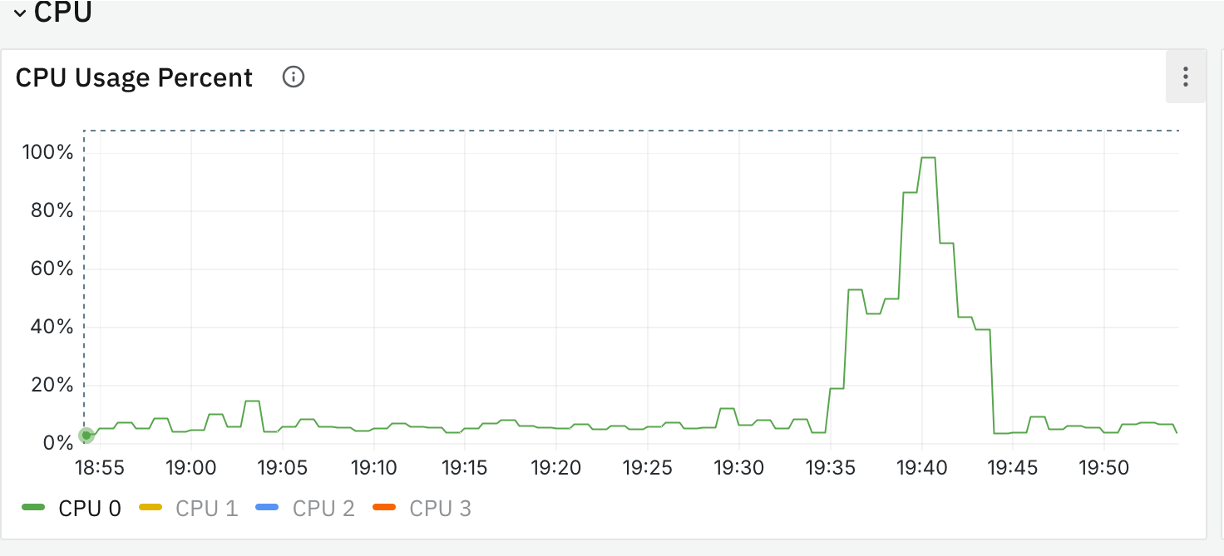}
\caption{Monitoring system for visualizing metrics}
\label{fig:ibm-monitoring}
\end{figure}

\begin{figure}[h]
\centering
\includegraphics[width=1\linewidth]{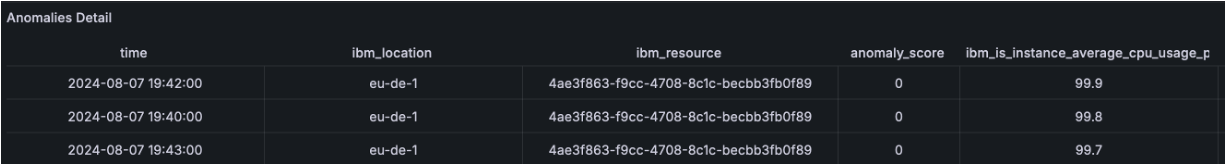}
\caption{Grafana Dashboard for visualizing the anomalies}
\label{fig:grafana}
\end{figure}

We calculate the mean and standard deviation of the array of reconstruction errors generated by the ReconstructAD algorithm. Subsequently, we employ the Chi-Squared distribution to calculate the p-values, which are returned as an anomaly score for each anomaly. Finally, we utilize a threshold value to determine whether an anomaly is significant enough to be flagged for further investigation or action. This threshold can be adjusted based on the desired trade-off between false positives and false negatives, depending on the specific requirements of the organization. Principal Component Analysis (PCA) is employed to identify the most influential metrics contributing to anomalies in multivariate anomaly detection scenarios.

\item \textbf{System Visualization}
We designed and developed a Grafana \cite{grafana_docs}-based dashboard for Site Reliability Engineers (SREs) to visualize and monitor anomalies in real-time, providing a centralized platform for incident detection and response. The Figures presented, Figures \ref{fig:ibm-monitoring} and \ref{fig:grafana}, provide valuable insights into the functionality of the anomaly detection system. They demonstrate its capability to identify spikes or anomalies and visually represent these on a dashboard in a tabular format. Figure \ref{fig:ibm-monitoring} shows a screenshot of our Monitoring system, highlighting the \textit{CPU Usage Percent} waveform with a potential spike. Figure \ref{fig:grafana} presents a Grafana dashboard screenshot of those spikes detected as anomalies in a VSI (Virtual Server Instance) at a specific timestamp, specifically illustrating the Sysdig metric \textit{ibm\_is\_instance\_average\_cpu\_usage\_percentage}. These visualizations enable users to quickly identify and respond to anomalies, thereby improving the overall efficiency and reliability of the cloud infrastructure.
\end{enumerate}

\section{Practical Scenarios: A Persona-Centric Approach}
In this section, we explore how various personas within the operations and development teams can benefit from our anomaly detection system.

\subsection{Operations SREs (Site Reliability Engineers)}
Currently, Operations SREs spend a significant amount of time manually monitoring various dashboards to identify issues in the infrastructure. In a region with approximately 20,000 VSIs (Virtual Server Instances), it is impractical to monitor each VSI individually for problems. Our anomaly detection system automates this process, substantially reducing the manual effort required. By flagging anomalies in real-time, the system enables SREs to focus on validating and addressing these issues, rather than sifting through multiple dashboards.

\subsection{Alert Developers}
Our system provides precise anomaly notifications, which can be leveraged to fine-tune existing rule-based alert systems\cite{10305844}. For instance, a rule-based alert might trigger if CPU utilization exceeds 80\%. However, our system can detect anomalies even if the utilization spikes from 60\% to 75\% briefly and then drops back to 60\%, which would not trigger the rule-based alert. This enables alert developers to adjust thresholds intelligently and create more effective alerting mechanisms, thereby improving the overall monitoring strategy.

\subsection{Service SREs (Site Reliability Engineers)}
Service SREs can utilize our anomaly detection system to monitor the performance and reliability of micro-services running in containers. By identifying performance bottlenecks and potential failures early, they can maintain service quality and improve the user experience. The system also facilitates proactive decision-making to mitigate issues that might not be captured using traditional rule-based alerting systems.

\section{The Anomaly Detection Service API}
\begin{figure}[!h]
\centering
\includegraphics[scale=0.83]{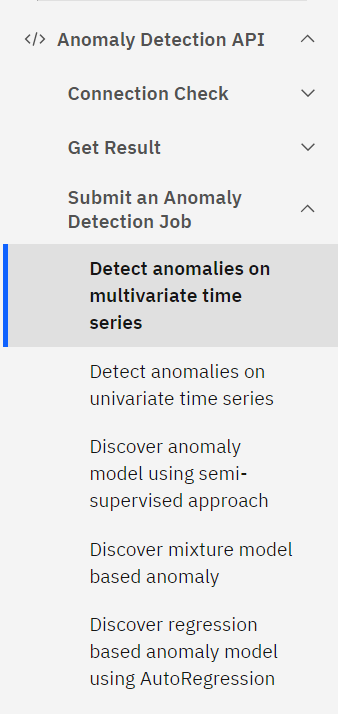} 
\caption{Anomaly Detection Service}
\label{endpoints}
\end{figure}

The anomaly detection model employed by our deployed system for SREs leverages our Anomaly Detection Service API, which we will detail in this section. Figure \ref{endpoints} illustrates the anomaly detection service. 

This service is an early attempt to onboard end user to use the time series anomaly detection API to explore the range of anomaly detection algorithms and obtain feedback. We explore anomaly detection as it is a application in many data-driven real-world applications such as IoT monitoring, web of things, asset management, and more, where there are rare ocurrences of failures where the behavior is out of the norm. We focus on time-series as it is the most commonly supported data format generated from sensors. By exposing a range of anomaly detection algorithms via web APIs, we achieve three key objectives:

\begin{enumerate}
\item Simplify the construction and execution of various types of anomaly pipelines/workflows
\item Protect the Intellectual Property (IP) related to core algorithm implementation by hiding the details of the code via a service
\item Auto-scale the system with a pay-as-you-go model, and enable the user with provisioning of the right resources
\end{enumerate}

The web based API Service discussed in this paper is deployed on a public cloud for a general audience (trial subscription only). We carefully designed an API for exploring many algorithms and workflows. Our key innovation for web-based API consists of the following features:

\begin{enumerate}
\item Five purpose-built anomaly workflows for jump-starting, including Univariate, Multi-Variate, Semi-Supervised, Regression-based, and Gaussian-Mixture-based (See Figure \ref{endpoints}). These workflows provide a solid foundation for users to quickly get started with anomaly detection.
\item Each purpose-built workflow offers sub-control over the types of models to run, allowing users to fine-tune their anomaly detection approach. For example, the Multi-variate Anomaly Detection endpoint provides five anomaly learning pipelines, along with a choice of 16 different algorithms and scoring options.
\item Support for on-demand execution environments (e.g., Setting M, L, etc.) to adjust development costs and optimize resource utilization.
\item The ability to pass input data via either payload or COS buckets, providing flexibility in data ingestion.
\item Robust job execution capabilities, supporting up to 2 hours of job execution and 100 service jobs simultaneously, ensuring that users can process large datasets efficiently.
\end{enumerate}

\subsection{Anomaly Detection Service Cheat Sheet}
As illustrated in Figure \ref{endpoints}, we have published five API endpoints for anomaly detection tasks at the time of writing this paper. These endpoints cater to a broad range of anomaly detection use cases originating from various IoT-based AI applications \cite{BELIS2024100091, ARISEKOLA2023104833, app11125320, CHATTERJEE2022100568}. Each endpoint is designed to train an AI model that can identify data points that significantly deviate from the normal pattern of the data sequence. Users have the flexibility to submit anomaly detection jobs for either univariate or multivariate time series data. Initially, we published two anomaly detection services for detecting anomalies from univariate and multivariate time series in an unsupervised fashion. Both endpoints support two distinct anomaly scenarios:

\begin{table*}[t!]
\centering
\caption{Anomaly Detection Service Cheat Sheet (Version 1.2.3). The tilde \textasciitilde \ symbol is used to denote shorthand notation.}

\label{platform}
\begin{tabular}{l|c|c|c|c|c}
\toprule
API arguments & \rot{Univariate} & \rot{Multi-variate} & \rot{Semi-supervised} & \rot{Regression-based} & \rot{Mixture-Model-based} \\ \hline
data\_file & \checkmark & \checkmark & \checkmark & \checkmark & \checkmark \\ \hline
time\_column & \checkmark & \checkmark & \checkmark & \checkmark & \checkmark \\ \hline
time\_format & \checkmark & \checkmark & \checkmark & \checkmark & \checkmark \\ \hline
target\_column(s) & \checkmark & \checkmark & \checkmark & \checkmark & \checkmark \\ \hline
label\_column &  &  & \checkmark & & \\ \hline
feature\_columns & & & & \checkmark & \\ \hline
prediction\_type & \checkmark & \checkmark & & & \\ \hline
recent\_data & \checkmark & \checkmark & & & \\ \hline
algorithm\_config & \checkmark & \checkmark & \checkmark & \checkmark & \checkmark \\ \hline
algorithm\_type & \checkmark & \checkmark & & & \\ \hline
anomaly\_estimator & \checkmark & \checkmark & & & \\ \hline
lookback\_window & \checkmark & \checkmark & & & \\ \hline
observation\_window & \checkmark & \checkmark & & & \\ \hline
labeling\_method & \checkmark & \checkmark & & & \\ \hline
labeling\_threshold & \checkmark & \checkmark & & & \\ \hline
train\_val\~\_test\_column & & & \checkmark & & \\ \hline
evaluation\_metrics & \checkmark & \checkmark & \checkmark & \checkmark & \checkmark \\ \hline
evaluation\_time & \checkmark & \checkmark & \checkmark & \checkmark & \checkmark \\ \hline
instance\_size & \checkmark & \checkmark & \checkmark & \checkmark & \checkmark \\ \hline
unsupervised\_fs\~ & \checkmark & \checkmark & \checkmark & \checkmark & \checkmark \\ \hline
train\_test\_split & & & & \checkmark & \checkmark \\ \hline
train\_cv\_split & & & & \checkmark & \checkmark \\
\bottomrule
\end{tabular}
\label{adtbl}
\end{table*}

The two anomaly detection scenarios supported by our API endpoints are:

\begin{enumerate}
    \item \textbf{Batch}. Detect anomalies within the given time series.
    \item \textbf{Stream}. Detect anomalies in the recent time points with respect to previously analyzed time series.
\end{enumerate}

These scenarios are useful in many IoT-driven AI applications that require monitoring the OSI PI tag time series. As our tool gains more developer attention, we have recently enabled support for semi-supervised, regression-based, and mixture-model-based anomaly detection capabilities (Version 1.2.2). We have observed that these endpoints are useful for building anomaly models for a wide variety of industrial assets, such as Wind Turbines, Oil Pumps, Switch Gear, Standby Generators, Blast Furnaces, Financial data, etc. For example, a regression-based anomaly detection model is common for Wind Turbines assets, where "Power Output" is predicted using "Wind Speed", "Wind Direction", and "Ambient Temperature".

Table \ref{adtbl} provides a developer cheat sheet for the Anomaly Detection Service. Each row represents the API arguments, and each column refers to the 5 service endpoints. The brief descriptions of the 5 service endpoints are as follows:

\begin{enumerate}
    \item \textbf{Univariate}. Univariate time series is the type of data that consists of observations on only a single characteristic or attribute. Many IoT assets are monitored by multiple metrics, and anomaly detection on these metric time series is an active interest.
        
    \item \textbf{Multi-variate}. Multivariate data is the type of data that consists of observations on more than one characteristic or attribute. This endpoint submits a job to the service that detects anomalies on multivariate time series data provided by users.
        
    \item \textbf{Regression-based}. Many anomaly detection algorithms are based on a regression model that captures the complex relationship between variables defining the system performance. This service examines the effectiveness of an automatic regression model selection procedure \cite{9378082} to build characteristic healthy behavior of IoT Assets using the input variables (feature\_columns) and output variable (target\_column).
        
    \item \textbf{Mixture-Model based}. This service endpoint aims to identify structures and modes in the operation of an asset. A user can feed into our model a historical record of an asset operating in a normal fashion. After we train our model on the input variables specified by feature columns, we can identify the different modes of normal operation and an expected range of readings from the IoT sensors. The service provides an automated way of discovering the best mixture model parameters. Once we have trained our model, we can apply it to a new record of data to identify the mode in which our asset is currently operating and flag whether or not our asset is in a previously unseen/anomalous mode of operation.

    \item \textbf{Semi-Supervised based}. Failures are comparatively rare events, so there is often a scarcity of tagged data for building supervised predictive models. In the absence of abundant tagged (supervised) failure data, semi-supervised services provide an intermediate solution where model training occurs only using normal data, whereas model selection and/or parameter tuning is conducted under the influence of the very limited failure instances. Users have the option of what anomaly model to deploy based on the type of failure they are interested in.

\end{enumerate}

For ease of use, the meaning of API arguments are almost similar across various anomaly services. We organize the API arguments into 4 broad categories:

\begin{enumerate}
    \item \textbf{Data argument}. Users need to annotate the input data with future column, time column, target columns, as well as provide necessary credentials if the location of the data is in the COS bucket;
    \item \textbf{Algorithm configuration argument.} Each service runs a series of anomaly detection models. Users can provide necessary input using algorithm config, algorithm type, and anomaly estimator, etc.
    \item \textbf{Evaluation setting argument.} The execution of the anomaly detection process is controlled by various parameters such as execution time limit, instance size in terms of CPU and memory, evaluation metric, etc.
    \item \textbf{Output argument.} The anomaly detection algorithm traditionally generates the real values anomaly score. The output-related arguments help users obtain the anomaly label (+1 for normal and -1 for abnormal) using various parameters such as labeling method, labeling threshold, etc. 
\end{enumerate}
\section{Anomaly Benchmark}
To demonstrate the quality of the underlying algorithm, we conducted an in-depth benchmark analysis. We opted to leverage the existing benchmark suite \cite{9458835}, which has been rigorously evaluated with the state-of-the-art framework DAEMON (Adversarial Autoencoder Anomaly Detection Interpretation). This choice was driven by the architecture's robust performance across diverse datasets. We used three datasets: SMD, MSL, and SMAP, for our analysis, and each dataset has multiple assets. Following benchmark literature, we trained an anomaly model for each asset separately and then generated anomaly scores on the test portion of the dataset. Once the scores were generated, we obtained the evaluation score (F1, Precision, and Recall) using the ground truth information available for the test dataset. As suggested in the original paper \cite{9458835}, we aggregated the results for each dataset. We used 14 algorithms exposed via endpoint 2 (detecting anomalies in multivariate time series) to generate the anomaly scores.

Table \ref{result} provides a summary of the F1 score for each dataset. The first row is the result reported in the original paper. It can be seen that the models from our suite of anomaly detection algorithms are competitive, if not better, than the stated results in more than one dataset. \texttt{DAEMON} seems to have the highest F1 for SMD data, although \texttt{MachineTranslation} and \texttt{GMM\_L0}, \texttt{GMM\_L1} are close.  \texttt{GMM\_L1} has the best F1 for MSL by a long way, and it is seen that a few other algorithms perform better than or very close to \texttt{DAEMON}.  \texttt{GMM\_L0} has the best result for SMAP, and \texttt{DeepAD} is a close second. Overall it  is seen that the scores are  varying but the anomaly models are very generalizable for small as well as large data. Our benchmark also validated the robustness of the API to tackle the variety of time series both from a performance (i.e. very large time series, long training times) and quality perspective.

\begin{table}[t!]
\small
\centering
\begin{tabular}{l|l|l|l}
\toprule
Pipeline\_Estimator & SMD & SMAP & MSL \\ \hline
\texttt{DAEMON}  & \textbf{0.963} & 0.91  &  0.953 \\ \hline
\texttt{DNN\_AutoEncoder}  & 0.862  & 0.647  & 0.829  \\ \hline
\texttt{CNN\_AutoEncoder}  & 0.582  & 0.596  & 0.610 \\ \hline
\texttt{Seq2seq\_AutoEncoder}  & 0.682  & 0.604  & 0.612 \\ \hline
\texttt{DNN\_VarAutoEncoder}  & 0.765 & 0.669  & 0.708 \\ \hline
\texttt{IsolationForest}  & 0.865 & 0.715  & 0.809 \\ \hline
\texttt{AnomalyEnsembler}  & 0.876  & 0.715  & 0.817 \\ \hline
\texttt{NSA}  & 0.559  & 0.675  & 0.651  \\ \hline
\texttt{NearestNeighbor}  & 0.812 & 0.713  & 0.650 \\ \hline
\texttt{PredAD}  & 0.903  & 0.851  & 0.934  \\ \hline
\texttt{DeepAD} & 0.936  & 0.970  & 0.934  \\ \hline
\texttt{GMM\_L1} & 0.947 & NA & \textbf{0.957}\\ \hline
\texttt{GMM\_L0} & 0.956 & \textbf{0.985} & 0.956 \\ \hline
\texttt{Covariance} & 0.741 & 0.589 & 0.682 \\ \hline
\texttt{MachineTranslation} & 0.946 & 0.884 & 0.868 \\ \hline
\end{tabular}
\caption{Benchmark Results: Average F1 score.}
\label{result}
\end{table}

\section{Application of Anomaly Detection in Cloud Monitoring and API Usage}

The API has been widely used by data scientists, business users, independent service vendors as well as researchers. Between Jan-2022 to till dates, 500k+ API calls are being made, with at-least 200 calls on a monthly basis. We have excluded any API call originating from the author of this paper as well as the service call from non-trial subscriptions. We now discuss an application build using service for monitoring cloud resources.

\section{API Adoption via API Hub}

\begin{figure}[h]
    \centering
    \includegraphics[width=1\linewidth]{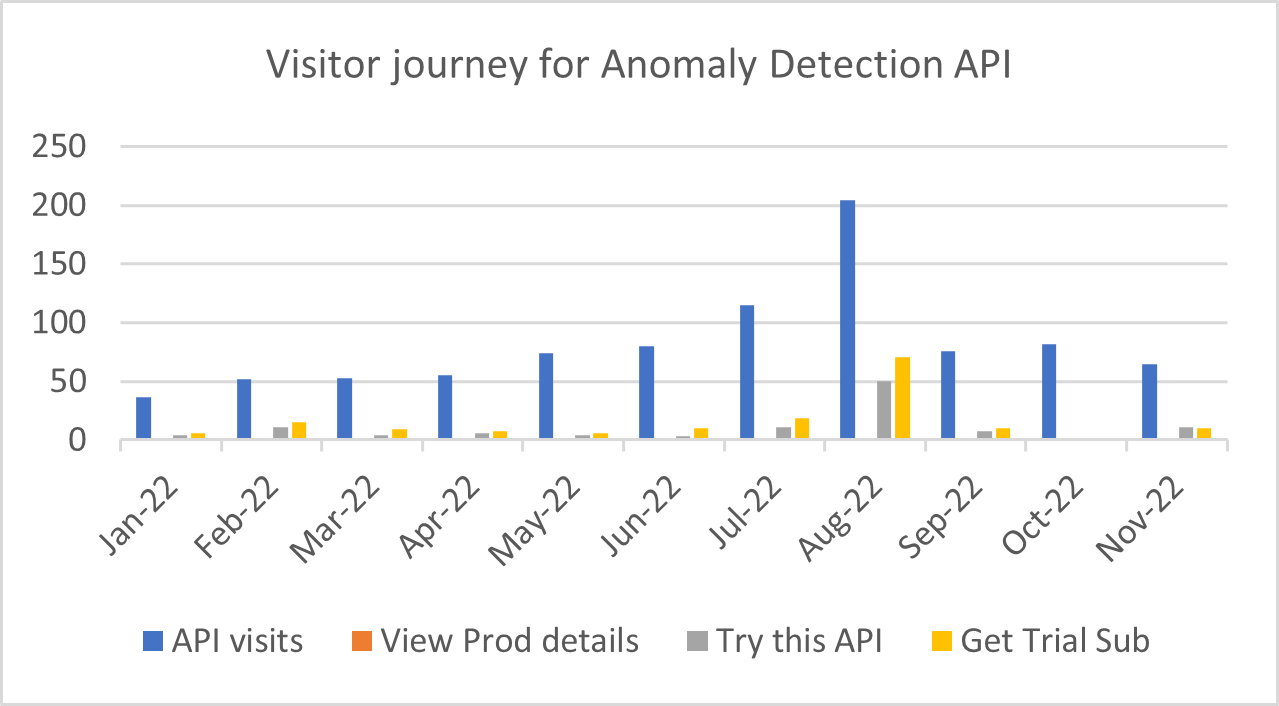}
    \caption{Continuous Visitor Journey (with KDD-2022 Anomaly Detection Tutorial with close to 200+ visits and 150+ participants}
    \label{fig:apiusage1}
\end{figure}

The Anomaly Detection Service @ IBM, developed by IBM Research\footnote{\url{https://developer.ibm.com/apis/catalog/ai4industry--anomaly-detection-product/}}, enables users to make API calls free of charge for detecting anomalies. In Figure \ref{fig:apiusage1}, we have captured important metrics over the year 2022 as an initial testing phase.

We interacted with nearly 300+ external practitioners over a course of one year via various conference forums (DASFAA-2022, ICDE-2022, KDD-2022, MLSys-2022, AAAI-2023) and demos (AAAI-2021). We developed an in-depth tutorial on building an anomaly model for time series data\footnote{\url{https://github.com/IBM/anomaly-detection-code-pattern/tree/main/tutorials}}. Our discussion with the community brought up many interesting questions:

\begin{enumerate}
    \item How can I train foundation models for Anomaly Detection?
    \item How can you be certain about your anomaly prediction?
    \item How do we generate actionable insights from anomaly models?
    \item How do I operationalize anomaly models? 
    \item What is an MLOps (Machine Learning Operations) for unsupervised models?
    \item How do we validate anomaly scores?
\end{enumerate}

\begin{figure}[h]
    \centering
    \includegraphics[width=1\linewidth]{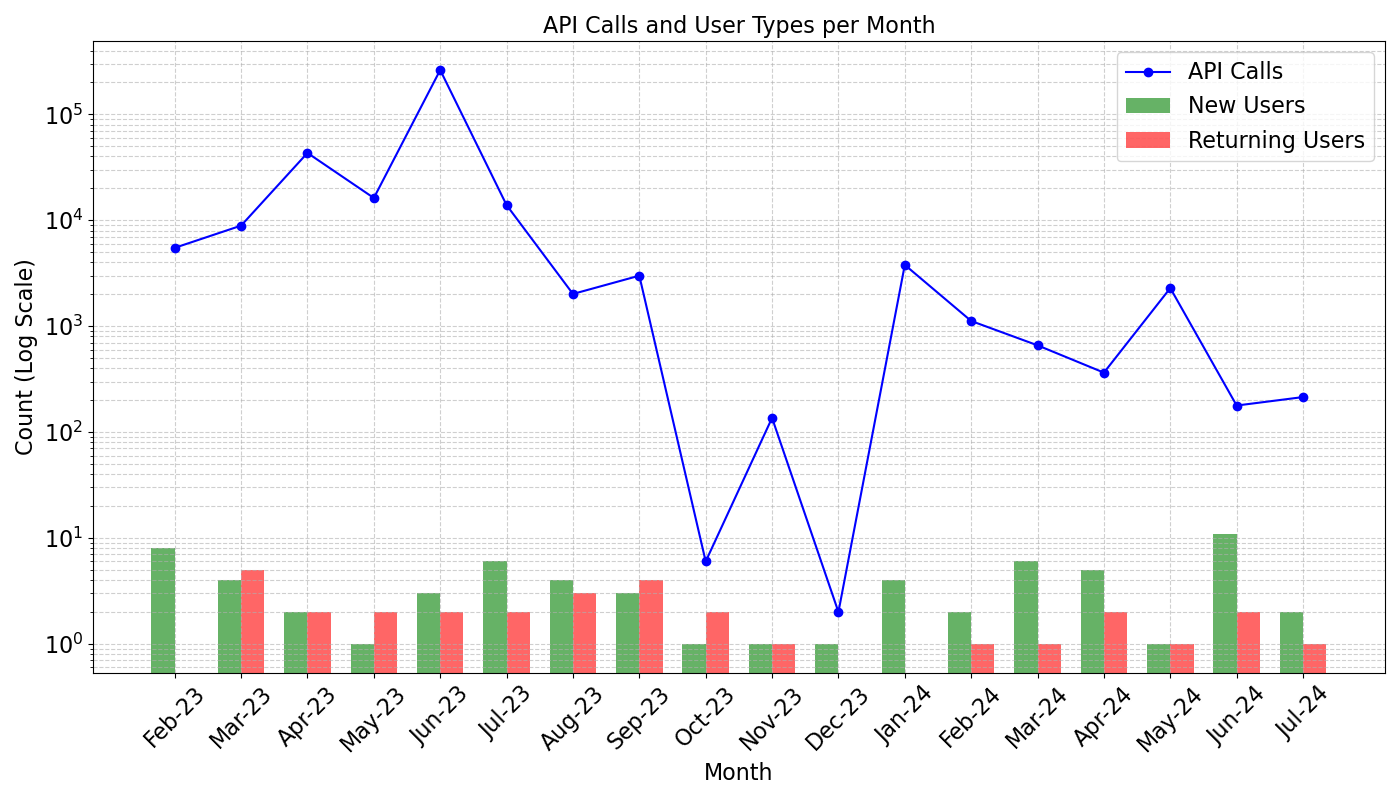}
    \caption{Combined plot of API calls (line) and user statistics (bars).}
    \label{fig:apiusage}
\end{figure}

We continue our effort to address the above feedback. The API Hub platform has a way to obtain statistics about how many users are getting on-boarded and using them in their day-to-day work. Although the service has been in place since 2021 and has enabled more than 500+ users, Figure \ref{fig:apiusage} displays a combined view of API activity and user statistics over time, from February 2023 to July 2024. The line plot represents the number of API calls made each month, while the grouped bar plots illustrate the counts of new and returning users.

The number of API calls exhibits significant variation, with a sharp peak in June 2023. To better visualize the data, the y-axis employs a logarithmic scale, which effectively handles the wide range of values. The number of new users fluctuates throughout the period, with notable increases in June 2023 and Jun 2024. Similarly, the count of returning users also fluctuates, with peaks in March 2023 and June 2024. The high volume of API calls in June 2023 indirectly suggests a scalable capability of the implementation, demonstrating its ability to handle a large number of requests. 

\section{Conclusion}
This paper introduces a scalable and generalizable Anomaly Detection system deployed on the cloud, which utilizes our Anomaly Detection API to augment the role of Site Reliability Engineers (SREs) in cloud infrastructure management. The system provides a thorough overview of anomaly detection capabilities, encompassing algorithms, decision functions, REST API design, and practical applications.
We also discuss how our Large Language Model (LLM)-assisted anomaly modeling approach effectively captures anomalous behaviors of various cloud infrastructure components by systematically leveraging pre-trained LLMs. This approach demonstrates its potential in enhancing cloud infrastructure resilience, reducing downtime, and facilitating root cause analysis. By utilizing our service, SREs can proactively identify potential issues before they escalate, thereby reducing downtime and improving response times to incidents, ultimately leading to an enhanced overall customer experience.
The deep learning-based Anomaly Detection pipeline presented in this paper is currently being explored for the utilization of time series foundation models, with a focus on leveraging the zero-shot capability of time series forecasting models. Through a comprehensive benchmark analysis, we demonstrate that our models are competitive with state-of-the-art results on several datasets.

\bibliography{sample-base}

\newpage
\appendix
\section{Appendix}
\subsection{LLM-assisted anomaly modeling example}
\label{llm_examp}
Example:
\begin{itemize}
    \item Server Failure Modes
    \begin{itemize}
        \item Power supply failure
        \begin{itemize}
            \item Behavior: Sudden spikes or sustained high power usage, indicating potential power supply issues or hardware failure
        \end{itemize}
        \item Firmware or software corruption
        \begin{itemize}
            \item Behavior: Sudden spikes or sustained high CPU usage, indicating potential resource starvation or software issues
        \end{itemize}
    \end{itemize}
    \item Database Failure Modes
    \begin{itemize}
        \item Too many Database Connections
        \begin{itemize}
            \item Behavior: Sudden spikes or sustained high database connections, indicating potential resource starvation or software issues
        \end{itemize}
        \item Too many Database Queries
        \begin{itemize}
            \item Behavior: Sudden spikes or sustained high database queries, indicating potential resource starvation or software issues
        \end{itemize}
    \end{itemize}
\end{itemize}

Figure \ref{casy-example-output} shows an example from the relevant metrics to monitor after the LLM has mapped it to the given variables names from the available cloud infrastructure metrics from our dataset.

\newpage
\begin{figure}[ht]
\begin{mybox}{enhanced, title=LLM-assisted anomaly modelling final response}
\small
Here are the metrics to use to build an anomaly model and the behavior to monitor for each metric:\\

\textbf{Server Metrics}
\begin{itemize}
    \item \textbf{CPU Usage} 
    \begin{itemize}
        \item $ibm\_is\_instance\_average\_cpu\_usage\_percentage$
        \item $ibm\_is\_instance\_cpu\_usage\_percentage$
    \end{itemize}
    \textbf{Behavior to monitor}: Sudden spikes or sustained high CPU usage, indicating potential resource starvation or software issues\\
    \textbf{...}
    \item \textbf{Disk Usage} 
    \begin{itemize}
        \item $ibm\_is\_instance\_volume\_read\_bytes$
        \item $ibm\_is\_instance\_volume\_write\_bytes$
    \end{itemize}
    \textbf{Behavior to monitor}: Sudden spikes or sustained high disk usage, indicating potential resource starvation or software issues\\
\end{itemize}
\textbf{...}\\

\textbf{Network Metrics}
\begin{itemize}
    \item \textbf{Network Traffic} 
    \begin{itemize}
        \item $ibm\_is\_instance\_network\_in\_bytes$
        \item $ibm\_is\_instance\_network\_out\_bytes$
    \end{itemize}
    \textbf{Behavior to monitor}: Sudden spikes or sustained high network traffic, indicating potential network congestion or software issues\\
    \textbf{...}
    \item \textbf{Network Errors} 
    \begin{itemize}
        \item $ibm\_is\_instance\_network\_in\_errors$
        \item $ibm\_is\_instance\_network\_out\_errors$
    \end{itemize}
    \textbf{Behavior to monitor}: Sudden spikes or sustained high network errors, indicating potential network issues or software problems\\
\end{itemize}
\end{mybox}
\caption{LLM-assisted anomaly modelling final response}
\label{casy-example-output}
\end{figure}

\end{document}